\documentclass[11pt]{article}
\usepackage[preprint]{acl}
\usepackage{times}
\usepackage{latexsym}
\usepackage[T1]{fontenc}
\usepackage[utf8]{inputenc}
\usepackage{microtype}
\usepackage{inconsolata}
\usepackage{graphicx}
\usepackage{algorithm}
\usepackage{algorithmic}
\usepackage{tabularx}
\usepackage{multirow}
\usepackage{booktabs}
\usepackage{amssymb} 
\usepackage{amsmath} 
\usepackage{makecell}
\usepackage[table]{xcolor}
\usepackage{colortbl}

\title{Import What You Need: Learning When and How to\\ Augment EHR Graphs with External Knowledge}

\author{
  Chen Chen$^{\dagger}$, 
  Mohsen Nayebi Kerdabadi$^{\dagger}$, 
  Dongjie Wang$^{\dagger}$,
  Mei Liu$^{\ddagger}$,
  Zijun Yao$^{\dagger}$\thanks{Corresponding author.}\\[6pt]
  $^{\dagger}$Electrical Engineering and Computer Science, University of Kansas, USA\\
  \texttt{\{chenchen, mohsen.nayebi, wangdongjie, zyao\}@ku.edu}\\[3pt]
  $^{\ddagger}$Health Outcomes and Biomedical Informatics, University of Florida, USA\\
  \texttt{mei.liu@ufl.edu}\\[3pt]
}

\newcommand{\modelname}{ReTA}

\begin{document}
\renewcommand{\thefootnote}{\fnsymbol{footnote}}
\maketitle
\renewcommand{\thefootnote}{\arabic{footnote}}
\setcounter{footnote}{0}

\begin{abstract}
Longitudinal prediction from electronic health records (EHRs) is limited by the sparsity and irregularity in patient trajectories, and knowledge augmentation with external knowledge graphs (KGs) offers a promising way to alleviate these issues.
However, KG expansion often follows predefined rules, with limited control over whether and how to augment each visit as a patient's state evolves.
We propose \textbf{\modelname{}}, a \underline{\textbf{Re}}inforcement learning-based dynamic \underline{\textbf{T}}opology \underline{\textbf{A}}ugmentation framework that casts KG import as a per-visit, budget-aware policy.
\modelname{} first constructs an offline refined pool of KG-grounded templates, then learns a policy to select one augment action per visit from three options: Soft Import, which enriches node features without modifying graph topology,
Hard Import, which grafts a compact KG subgraph onto the visit graph to create message-passing shortcuts,
and Skip, which leaves the visit unaugmented when the base encoder is already confident.
To stabilize learning, \modelname{} employs a decoupled encoder that processes semantic and structural signals in separate channels and fuses them via adaptive gating.
Experiments on MIMIC-III and MIMIC-IV across diagnosis prediction, mortality, and readmission show that \modelname{} consistently outperforms strong baselines while remaining efficient, with larger gains under sparse supervision. 
\modelname{} transfers across datasets and knowledge graphs and yields interpretable augmentation patterns.
\end{abstract}

\section{Introduction}
\label{sec:introduction}

Massive electronic health records (EHRs) have enabled a wide range of healthcare prediction tasks, from mortality risk estimation to disease progression modeling~\cite{choi2016retain,ma2017dipole,choi2017gram,luo2020hitanet}.
However, patient-level trajectories (e.g., medical codes over visits) in EHRs are often sparse and irregular~\cite{rasmy2021med}, making it difficult for data-driven approaches to capture complex clinical patterns such as comorbidities, complication cascades, and organ-system interactions.
Established knowledge graphs (KGs) in healthcare, such as PrimeKG~\cite{chandak2023building}, offer clinically validated concepts and relations that complement raw visit data.
Topology augmentation, which injects KG-derived structure into EHR visit graphs, has therefore emerged as a promising strategy for learning knowledge-augmented patient representations~\cite{liu2020k}.

Despite their promise, existing KG-enhanced topology augmentation methods commonly exhibit three limitations.
\textbf{Static}. Code-level expansion commonly follows predefined KG neighborhoods.
The resulting graph can be patient-specific, yet the expansion rule may still overlook how the value of additional knowledge changes from one visit to the next.
This can introduce weakly related concepts and increase message-passing cost~\cite{xu2023seqcare, jiang2024graphcare, zhu2024emerge}.
\textbf{Unbudgeted}. Most approaches do not enforce a resource budget for KG augmentation.
Without budget control, expensive strategies (e.g., hard structural injection) can dominate, while lighter alternatives or outright abstention are not systematically leveraged, leading to over-augmentation that amplifies noise~\cite{liu2024graph, zhang2024graph}.
\textbf{Single-pass}. Augmentation is commonly performed in a single shot for an entire trajectory, rather than making adaptive decisions at each visit.
Injected knowledge may therefore not reflect evolving patient states.
For example, a patient transitioning from chronic diabetes management to acute sepsis needs different relational context at each stage, but single-pass methods apply the same expansion throughout~\cite{ye2021medpath, wang2025colacare}.

In this work, we propose \modelname{}\footnote{Code: \url{https://github.com/ChenC2002/ReTA/}.}, a dynamic, budgeted, and adaptive framework that performs visit-level KG augmentation based on each patient's evolving history. 
\modelname{} formulates KG injection along patient progressions as a sequential decision problem in three steps.

\begin{figure}[t]
    \centering
    \includegraphics[width=\linewidth]{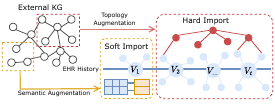}
    \vspace{-10pt}
    \caption{Two augmentation modes for visit-level KG injection. \textbf{Soft Import} enriches EHR concepts via semantic embeddings without modifying visit graph structure, while \textbf{Hard Import} augments topology by adding KG-derived nodes and edges.}
    \label{fig:Editing}
    \vspace{-10pt}
\end{figure}

\textbf{Step 1}. We construct a quality-filtered knowledge pool offline.
Each medical concept is distilled via an LLM into a compact template (a semantic definition paired with a structural cascade), grounded against PrimeKG, and clustered to remove redundancy.
At each visit, the top-$K$ most relevant templates are retrieved by combining code-level similarity with trajectory-level context.

\textbf{Step 2}. We introduce a budget-aware augmentation action space (Figure~\ref{fig:Editing}).
\textit{Soft Import} enriches node features without modifying graph topology.
\textit{Hard Import} grafts a compact KG subgraph onto the visit graph to create new message-passing paths.
\textit{Skip} leaves the visit unaugmented when the base encoder is already confident.
A reinforcement learning policy selects one action per visit, balancing predictive gain against augmentation cost across the patient trajectory.

\textbf{Step 3}. Because Soft Import and Hard Import produce heterogeneous signals, we decouple encoding into a semantic channel and a structural channel, fused by an adaptive gate that selectively combines complementary evidence.

We evaluate \modelname{} on MIMIC-III~\cite{johnson2016mimic} and MIMIC-IV~\cite{johnson2023mimic} across three clinical tasks (diagnosis prediction, in-hospital mortality, and 30-day readmission).
\modelname{} outperforms strong baselines across all evaluated tasks at a lower inference cost.
Ablation studies (\S\ref{sec:ablation}) disentangle the effects of the knowledge pool and the policy's decisions, demonstrating that dynamically deciding whether and how to augment each visit outperforms conventional augmentation strategies in both accuracy and latency.
Moreover, the augmentation patterns are interpretable at the visit level, with performance gains widening as supervision becomes sparser.
Finally, \modelname{} transfers across clinical coding systems. When trained on MIMIC-IV and evaluated on MIMIC-III, it achieves a higher AUPRC than the strongest MIMIC-III-trained baseline. 

\section{Methodology}
\label{sec:method}

\begin{figure*}[t]
  \centering
  \includegraphics[width=\linewidth]{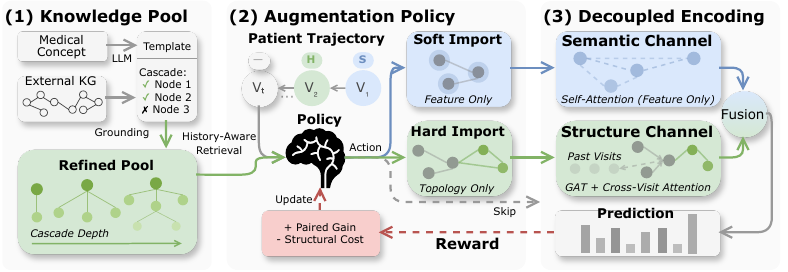}
  \caption{Overview of \modelname{}. \textbf{(1)} Each medical concept is distilled into a KG-grounded template and clustered into a refined pool. \textbf{(2)} The policy selects one action per visit from the retrieved candidates. \textbf{(3)} The encoder processes semantic and structural signals in separate channels before fusing them for prediction.}
  \label{fig1}
  \vspace{-5pt}
\end{figure*}

\subsection{Problem Setup}
\label{sec:problem_formulation}

\paragraph{Prediction task.}
Let $\mathcal{C}_{\mathrm{dx}}$ denote the set of all unique diagnosis codes in ICD-9/10.
A patient trajectory $\mathbf{V}=\{V_1,\dots,V_T\}$ is a chronological sequence of hospital visits, where each visit $V_t\subseteq\mathcal{C}_{\mathrm{dx}}$ records the diagnosis codes observed at time $t$.
We predict next-visit diagnoses at the granularity of Clinical Classifications Software (CCS) categories, which group fine-grained ICD codes into clinically meaningful phenotypes and keep the label space consistent across ICD-9 and ICD-10 cohorts.
Let $\mathcal{A}$ denote the CCS label space.
Given history $V_{1:t}$, the model outputs $\hat{\mathbf{y}}_{t+1}\in[0,1]^{|\mathcal{A}|}$, a probability vector over CCS categories for visit $t{+}1$.
For binary clinical outcomes (mortality, readmission), we replace the multi-label CCS head with a single sigmoid output and train with binary cross-entropy, keeping all other components identical.

\paragraph{Visit graphs.}
For each visit $V_t$ we build a graph $G_t=(\mathcal{V}_t,\mathcal{E}_t)$ whose nodes are the observed ICD codes together with their CCS ancestors up to $h$ hierarchy levels, and whose edges follow the CCS hierarchy among these nodes, treated as undirected for message passing (Appendix~\ref{app:B1_topology}).
We augment these visit graphs with knowledge from PrimeKG~\cite{chandak2023building}, a precision-medicine knowledge graph that integrates over twenty biomedical resources with rich pathophysiological and disease-complication relations.
The bridging between ICD/CCS and PrimeKG vocabularies is described in \S\ref{sec:distillation}.
Figure~\ref{fig1} illustrates the full pipeline.

\subsection{Knowledge Pool}
\label{sec:prompt_pool}

Per-visit augmentation is more efficient when it draws from a pre-built, bounded candidate set than when it expands an open-ended KG neighborhood on the fly.
We construct a pool of reusable knowledge templates offline, where each template pairs a semantic summary with a compact subgraph.

\subsubsection{LLM Distillation and KG Grounding}
\label{sec:distillation}
For each ICD code and its mapped CCS category, we prompt an LLM with the concept's canonical textual description and ask it to return a one-sentence definition describing the pathology and a clinical cascade listing downstream complications or comorbidities.
We adapt the cascade length to each concept's neighborhood density in PrimeKG. Concepts with few verified KG neighbors receive longer cascades (up to five items) to compensate for sparse relational context, while concepts with dense neighborhoods receive shorter ones (as few as one item).
For diabetes, which has moderate KG coverage, a typical cascade might list retinopathy, neuropathy, and nephropathy.
No patient-level context is provided, and each concept is processed independently with fixed decoding parameters (prompt template in Appendix~\ref{app:B3_distill}).

We ground each mention to a standardized biomedical identifier by exact matching against ICD, CCS, and PrimeKG vocabularies, falling back to cosine similarity between $\ell_2$-normalized name embeddings above a threshold $\tau_{\mathrm{map}}$~\cite{liu2021self,sung2020biomedical,neumann2019scispacy}.
This same normalization bridges ICD and CCS concepts to PrimeKG nodes when constructing visit graphs (\S\ref{sec:problem_formulation}).
For each concept, we then retrieve its PrimeKG neighborhood up to $k$ hops and retain only entities and links verifiable against the KG~\cite{soldaini2016quickumls}.
All filtering uses only external resources and fixed code descriptions.
Apart from the diagnosis vocabulary, pool construction touches no patient trajectories, outcomes, splits, or cohort statistics (thresholds and diagnostics in Appendix~\ref{app:grounding_details} and~\ref{app:pool_diagnostics}).

\subsubsection{Template Clustering}
\label{sec:clustering}

We compress and de-duplicate the grounded outputs by clustering.
Each concept text, consisting of its definition concatenated with its cascade, is embedded with ClinicalBERT~\cite{alsentzer2019publicly} and grouped via agglomerative clustering under cosine distance with cut threshold $\tau$.
For each resulting cluster we compute a centroid, project it into the model embedding space, and $\ell_2$-normalize it to obtain a template vector $\mathbf{p}_k\in\mathbb{R}^d$ (details in Appendix~\ref{app:B2_action_schema}).
The global pool $\mathcal{P}_{\mathrm{global}}=\{\mathbf{p}_k\}_{k=1}^M$ contains $M$ templates.
Because cascade lengths vary by concept, templates differ in subgraph size, which the augmentation policy accounts for when selecting actions (\S\ref{sec:rl_tuning}).

\subsubsection{History-Aware Retrieval}
\label{sec:retrieval}

At each visit $V_t$, we retrieve a candidate subpool from $\mathcal{P}_{\mathrm{global}}$ using both the current visit and the patient's accumulated history.
Let $s_t\in\mathbb{R}^{d_s}$ denote the patient state defined in \S\ref{sec:rl_tuning}, computed before augmentation.
We project this state into the template embedding space as $\tilde{s}_t=\mathbf{W}_{\mathrm{ret}}s_t$, where $\mathbf{W}_{\mathrm{ret}}\in\mathbb{R}^{d\times d_s}$.
Each template is scored using code-level and trajectory-level similarity.
\begin{align}
\mathrm{Score}(V_t,s_t,\mathbf{p}_k)
  ={} & (1{-}\alpha)\max_{c_i\in V_t}
        \cos(\mathbf{e}_{c_i},\mathbf{p}_k)
        \nonumber\\
      & + \alpha\,\cos(\tilde{s}_t,\mathbf{p}_k).
\label{Eq:score}
\end{align}
Here, $\mathbf{e}_c=\mathbf{W}_{\mathrm{code}}f_{\mathrm{text}}(\mathrm{desc}(c))\in\mathbb{R}^d$, where $\mathrm{desc}(c)$ is the canonical description of code $c$, $f_{\mathrm{text}}$ is the frozen ClinicalBERT encoder~\cite{alsentzer2019publicly}, and $\mathbf{W}_{\mathrm{code}}$ is a shared trainable projection. 
These representations serve as code-node features and code-level retrieval embeddings. 
The parameter $\alpha$ controls the weight of the trajectory signal.
We retain the top-$K$ templates by this score as the candidate subpool $\mathcal{P}^{(t)}_{\mathrm{sub}}$.

\subsection{Augmentation Policy}
\label{sec:rl_tuning}

Augmentation decisions along a trajectory are interdependent, since early actions reshape intermediate representations and constrain what is beneficial at later visits.
We make these decisions visit by visit in chronological order, formulating the process as a sequential decision problem trained with reinforcement learning.
Patient trajectories are short (median 2 visits, 75th percentile 3), so the policy performs short-horizon, cost-aware routing rather than long-horizon planning.

\subsubsection{State and Actions}

The state $s_t$ summarizes the patient history up to visit $t$.
Let $\mathbf{v}_t\in\mathbb{R}^d$ be the encoder's summary of visit $V_t$ (\S\ref{sec:decoupled_gat}) on the base graph before augmentation.
We compress past visits with a GRU~\cite{chung2014empirical} to form $s_t = \mathrm{GRU}(\mathbf{v}_{1:t-1}) \oplus \mathbf{v}_t$, where $\oplus$ denotes concatenation.

The discrete action space consists of $2K$ template-mode pairs plus a single Skip action, $\mathcal{O}^{(t)} = \big(\mathcal{P}^{(t)}_{\mathrm{sub}} \times \{0,1\}\big) \cup \{\texttt{Skip}\}$, giving $2K{+}1$ actions.
Each augmentation action $a_t=(\mathbf{p}_k,\delta)$ specifies which template to use and how to apply it.
The Skip action leaves the visit unaugmented.
A lightweight MLP policy $\pi_\theta(a_t\mid s_t)$ outputs a categorical distribution over these actions.
With one action per visit, the policy chooses whether to augment, which template to use, and whether to apply Soft or the costlier Hard Import.

We append $u_t=1-\max(\hat{y}^{\mathrm{base}}_{t+1})$, computed by the frozen Stage~1 encoder, to $s_t$. For diagnosis prediction, $u_t$ is the complement of the largest predicted label probability. For binary outcomes, it reduces to $1-p^{\mathrm{base}}_{t+1}$, giving the policy the predicted negative-class probability.

\paragraph{Soft Import.}
We map the selected template vector through a shared two-layer MLP $\phi$ and broadcast the resulting offset to every code in the visit, scaled by a strength parameter $\xi$,
\begin{equation}
\mathbf{X}'_t
  =\mathbf{X}_t
   +\xi\big(\mathbf{1}_{|V_t|}\otimes \phi(\mathbf{p}_k)\big),
\label{Eq:soft}
\end{equation}
where $\mathbf{X}_t\in\mathbb{R}^{|V_t|\times d}$ collects the node features of the original visit codes.
The offset is uniform because the template provides visit-level clinical context. Code-level differentiation happens in the downstream attention layers.

\paragraph{Hard Import.}
The template's compact subgraph $g_{\mathbf{p}_k}=(V_{\mathbf{p}_k},E_{\mathbf{p}_k})$ is grafted onto the visit graph to form $G'_t = (\mathcal{V}_t \cup V_{\mathbf{p}_k},\, \mathcal{E}_t \cup E_{\mathbf{p}_k})$, adding new relational paths for message passing between codes that share no direct edge.
Node features remain at their base values, so any gain from Hard Import is attributable to the added structure rather than to feature enrichment.

\subsubsection{Reward}

To isolate each augmentation's incremental benefit from the baseline difficulty of each visit, we compare the augmented input $G^{(t)}_{\mathrm{edit}}$ against its unaugmented counterpart $G^{(t)}_{\mathrm{raw}}$ under the current encoder on the same mini-batch with dropout disabled:
\begin{align}
r_t &= \lambda_1\Big(
       \mathcal{L}_{CE}\big(G^{(t)}_{\mathrm{raw}}\big)
      -\mathcal{L}_{CE}\big(G^{(t)}_{\mathrm{edit}}\big)
       \Big) \nonumber\\
    &\quad -\lambda_2\,\mathbb{I}[\delta{=}1]\,
       \frac{|V_{\mathbf{p}_k}|}{|\mathcal{V}_t|},
\label{Eq:reward}
\end{align}
where $\mathcal{L}_{CE}$ is the cross-entropy loss for predicting $y_{t+1}$, $\lambda_1$ weights prediction improvement, and $\lambda_2$ penalizes Hard Import in proportion to the added nodes.
Under Skip, raw and augmented inputs are identical, so $r_t = 0$ by construction.

To sharpen credit assignment, we maintain a running utility $\bar{r}_k$ for each template $k$, updated as $\bar{r}_k \leftarrow \gamma_r\, \bar{r}_k + (1{-}\gamma_r)\, r_t$ whenever template $k$ is selected.
The utilities of all $K$ candidates are concatenated with $s_t$ to form the policy input, giving the policy a memory of which templates have been useful in similar contexts.

We optimize the policy with REINFORCE~\cite{zhang2021sample} using a running-mean baseline to reduce variance.
With such short trajectories and a lightweight MLP policy, critic-based methods are unnecessary.

\subsection{Decoupled Encoding}
\label{sec:decoupled_gat}

Soft Import modifies node features while preserving graph topology.
Hard Import adds nodes and edges while preserving features.
Encoding both in a single channel would conflate these signals, making it difficult for the encoder to attribute gradient updates to the correct source.
We process them in two parallel channels and fuse the outputs with a learned gate.

\subsubsection{Semantic Channel}
This channel captures within-visit context from node features alone.
We apply multi-head self-attention over the original visit codes using their post-augment features $\mathbf{X}'_t$ to produce semantic representations $\{h^{\mathrm{sem}}_i\}_{i\in V_t}$.
Nodes added by Hard Import are excluded so that this channel reflects only observed codes.

\subsubsection{Structure Channel}
This channel runs message passing on the full augmented graph $G'_t$, including any nodes and edges added by Hard Import.
A graph-attention layer~\cite{velivckovic2018graph} aggregates neighbor information to produce within-visit structural representations $h^{\mathrm{intra}}_i$ for each node $i$.
To incorporate longitudinal context, each node also attends over past visit embeddings $\{h^{(t')}_G\}_{t'<t}$ via dot-product attention, yielding an across-visit summary $h^{\mathrm{inter}}_i$.
The two signals are combined with a sigmoid gate into $h^{\mathrm{struct}}_i = \mu_i\, h^{\mathrm{intra}}_i + (1{-}\mu_i)\, h^{\mathrm{inter}}_i$, where $\mu_i = \sigma(W_{\mathrm{fuse}}[h^{\mathrm{intra}}_i \| h^{\mathrm{inter}}_i] + b_{\mathrm{fuse}})$.

\subsubsection{Fusion}
We fuse the semantic and structural views with a second sigmoid gate $\beta_i = \sigma(W_{\mathrm{gate}}[h^{\mathrm{sem}}_i\|h^{\mathrm{struct}}_i] + b_{\mathrm{gate}})$, giving the final node representation $z_i = \beta_i\, h^{\mathrm{sem}}_i + (1{-}\beta_i)\, h^{\mathrm{struct}}_i$.
The visit embedding $h_G^{(t)}=\sum_{i\in V_t} z_i$ sums over original visit codes only, so injected nodes contribute through message passing but do not directly enter the prediction.

\subsection{Training}
\label{sec:training}

Training proceeds in two stages with the combined objective $\mathcal{L}_{\mathrm{total}}=\mathcal{L}_{\mathrm{task}}-\eta\,\mathcal{J}^{\mathrm{RL}}$, where $\mathcal{J}^{\mathrm{RL}}$ is the REINFORCE objective.

Stage~1 trains the encoder on $\mathcal{L}_{\mathrm{task}}$ while stochastically applying Soft Import, Hard Import, or Skip, producing a warm-started encoder that can stably process all three action types.
The encoder snapshot at the end of Stage~1 is frozen and supplies the base prediction $\hat{y}^{\mathrm{base}}_{t+1}$ used to compute the policy input $u_t$.

Stage~2 optimizes the augmentation policy under the paired reward while continuing to refine the encoder on policy-augmented visits, so that the encoder and policy co-adapt.
The full curriculum is given in Algorithm~\ref{alg:training}, with complexity analysis in Appendix~\ref{app:implementation}.

\section{Experiments}
\label{sec:experiments}

\begin{table*}[t]
  \centering
  \caption{Diagnosis prediction (mean$\pm$std, 5 seeds). GPT-4o results are averaged over five generation runs. Cross-Dataset rows are labeled train$\,\rightarrow\,$test. The best baseline is \underline{underlined}.}
  \label{tab:main_results}
  \vspace{-2mm}
  \renewcommand{\arraystretch}{1.15}
  \resizebox{\textwidth}{!}{%
  \small
  \setlength{\tabcolsep}{10pt}
  \begin{tabular}{ll ccc ccc}

    \toprule
    \multicolumn{2}{c}{} & \multicolumn{3}{c}{\textbf{MIMIC-III (\%)}}
      & \multicolumn{3}{c}{\textbf{MIMIC-IV (\%)}} \\
    \cmidrule(lr){3-5}\cmidrule(lr){6-8}
    \textbf{Category} & \textbf{Model}
      & AUPRC & Micro-F1 & Acc@20
      & AUPRC & Micro-F1 & Acc@20 \\
    \cmidrule(l){1-8}

    \multirow{2}{*}{\makecell[l]{\textit{LLM-}\\\textit{Based}}}
      & GPT-4o
      & 21.77{\scriptsize$\pm$0.35} & 18.94{\scriptsize$\pm$1.54} & 33.74{\scriptsize$\pm$0.33}
      & 22.05{\scriptsize$\pm$0.56} & 18.43{\scriptsize$\pm$0.88} & 33.88{\scriptsize$\pm$0.39} \\
      & \;\;+ PrimeKG
      & 25.55{\scriptsize$\pm$0.43} & 19.93{\scriptsize$\pm$1.68} & 36.53{\scriptsize$\pm$0.26}
      & 24.46{\scriptsize$\pm$0.56} & 20.30{\scriptsize$\pm$1.82} & 36.26{\scriptsize$\pm$0.31} \\
    \cmidrule(l){1-8}
    \multirow{2}{*}{\makecell[l]{\textit{Sequence}\\\textit{Models}}}
      & Transformer
      & 29.83{\scriptsize$\pm$0.61} & 24.59{\scriptsize$\pm$0.91} & 39.90{\scriptsize$\pm$0.46}
      & 29.25{\scriptsize$\pm$0.73} & 25.50{\scriptsize$\pm$1.17} & 39.48{\scriptsize$\pm$0.60} \\
      & RETAIN
      & 27.92{\scriptsize$\pm$0.68} & 23.02{\scriptsize$\pm$1.28} & 37.99{\scriptsize$\pm$0.44}
      & 28.22{\scriptsize$\pm$0.71} & 22.39{\scriptsize$\pm$1.77} & 38.63{\scriptsize$\pm$0.36} \\
    \cmidrule(l){1-8}
    \multirow{2}{*}{\makecell[l]{\textit{Ontology}\\\textit{Encoders}}}
      & HAP
      & 27.82{\scriptsize$\pm$0.42} & 23.83{\scriptsize$\pm$1.16} & 38.96{\scriptsize$\pm$0.43}
      & 31.75{\scriptsize$\pm$0.45} & 25.92{\scriptsize$\pm$1.19} & 41.18{\scriptsize$\pm$0.20} \\
      & G-BERT
      & 29.10{\scriptsize$\pm$0.39} & 23.70{\scriptsize$\pm$0.44} & 39.70{\scriptsize$\pm$0.42}
      & 32.95{\scriptsize$\pm$0.41} & 27.12{\scriptsize$\pm$0.94} & 42.56{\scriptsize$\pm$0.29} \\
    \cmidrule(l){1-8}
    \multirow{2}{*}{\makecell[l]{\textit{Structure}\\\textit{Learners}}}
      & GRAM
      & 27.24{\scriptsize$\pm$0.59} & 24.72{\scriptsize$\pm$1.33} & 38.40{\scriptsize$\pm$0.54}
      & 30.57{\scriptsize$\pm$0.26} & 26.08{\scriptsize$\pm$1.63} & 40.53{\scriptsize$\pm$0.49} \\
      & SeqCare
      & 30.81{\scriptsize$\pm$0.45} & 26.42{\scriptsize$\pm$1.46} & 40.75{\scriptsize$\pm$0.34}
      & 31.01{\scriptsize$\pm$0.51} & 27.63{\scriptsize$\pm$1.36} & 42.50{\scriptsize$\pm$0.67} \\
    \cmidrule(l){1-8}
    \multirow{3}{*}{\makecell[l]{\textit{KG-}\\\textit{Augmented}}}
      & GraphCare
      & 31.22{\scriptsize$\pm$0.59} & 26.17{\scriptsize$\pm$1.87} & \underline{42.51{\scriptsize$\pm$0.41}}
      & 31.49{\scriptsize$\pm$0.95} & 28.52{\scriptsize$\pm$0.85} & 42.82{\scriptsize$\pm$0.58} \\
      & RAM-EHR
      & 31.95{\scriptsize$\pm$0.59} & \underline{28.66{\scriptsize$\pm$1.33}} & 41.94{\scriptsize$\pm$0.46}
      & 32.41{\scriptsize$\pm$0.35} & \underline{32.38{\scriptsize$\pm$1.22}} & 44.40{\scriptsize$\pm$0.46} \\
      & KARE
      & \underline{32.37{\scriptsize$\pm$0.61}} & 28.26{\scriptsize$\pm$1.23} & 42.06{\scriptsize$\pm$0.53}
      & \underline{33.14{\scriptsize$\pm$0.35}} & 31.19{\scriptsize$\pm$1.42} & \underline{45.20{\scriptsize$\pm$0.42}} \\
    \cmidrule(l){1-8} 
    \multirow{2}{*}{\makecell[l]{\textbf{\textit{Ours}}\\\textbf{\textit{(ReTA)}}}}
      & \cellcolor[gray]{0.90} UMLS
      & \cellcolor[gray]{0.90} 33.65{\scriptsize$\pm$0.46}
      & \cellcolor[gray]{0.90} 29.25{\scriptsize$\pm$1.86}
      & \cellcolor[gray]{0.90} 43.86{\scriptsize$\pm$0.62}
      & \cellcolor[gray]{0.90} 34.28{\scriptsize$\pm$0.64}
      & \cellcolor[gray]{0.90} 31.03{\scriptsize$\pm$1.19}
      & \cellcolor[gray]{0.90} 45.79{\scriptsize$\pm$0.55} \\
      & \cellcolor[gray]{0.90} \textbf{PrimeKG}
      & \cellcolor[gray]{0.90} \textbf{34.18{\scriptsize$\pm$0.23}}
      & \cellcolor[gray]{0.90} \textbf{31.23{\scriptsize$\pm$1.76}}
      & \cellcolor[gray]{0.90} \textbf{44.14{\scriptsize$\pm$0.14}}
      & \cellcolor[gray]{0.90} \textbf{35.22{\scriptsize$\pm$0.41}}
      & \cellcolor[gray]{0.90} \textbf{32.69{\scriptsize$\pm$1.50}}
      & \cellcolor[gray]{0.90} \textbf{46.73{\scriptsize$\pm$0.44}} \\
    \midrule
    \midrule

    & & \multicolumn{3}{c}{\textbf{MIMIC-IV $\rightarrow$ MIMIC-III}}
    & \multicolumn{3}{c}{\textbf{MIMIC-III $\rightarrow$ MIMIC-IV}} \\
    \cmidrule(lr){3-5}\cmidrule(lr){6-8}
    \multirow{4}{*}{\makecell[l]{\textbf{Cross-}\\\textbf{Dataset}}}
      & GraphCare
      & 28.76{\scriptsize$\pm$0.40} & 24.26{\scriptsize$\pm$1.00} & 37.79{\scriptsize$\pm$0.32}
      & 26.01{\scriptsize$\pm$0.33} & 22.63{\scriptsize$\pm$0.97} & 36.56{\scriptsize$\pm$0.39} \\
      & RAM-EHR
      & 29.10{\scriptsize$\pm$0.64} & \underline{26.64{\scriptsize$\pm$0.17}} & \underline{38.78{\scriptsize$\pm$0.45}}
      & 27.78{\scriptsize$\pm$0.44} & \underline{26.45{\scriptsize$\pm$0.72}} & 38.09{\scriptsize$\pm$0.37} \\
      & KARE
      & \underline{29.33{\scriptsize$\pm$0.26}} & 25.79{\scriptsize$\pm$2.03} & 38.12{\scriptsize$\pm$0.17}
      & \underline{27.87{\scriptsize$\pm$0.59}} & 26.33{\scriptsize$\pm$1.37} & \underline{38.19{\scriptsize$\pm$0.34}} \\
      & \cellcolor[gray]{0.90} \textbf{\modelname{}}
      & \cellcolor[gray]{0.90} \textbf{33.09{\scriptsize$\pm$0.15}}
      & \cellcolor[gray]{0.90} \textbf{28.99{\scriptsize$\pm$1.71}}
      & \cellcolor[gray]{0.90} \textbf{42.46{\scriptsize$\pm$0.23}}
      & \cellcolor[gray]{0.90} \textbf{32.08{\scriptsize$\pm$0.86}}
      & \cellcolor[gray]{0.90} \textbf{29.29{\scriptsize$\pm$0.62}}
      & \cellcolor[gray]{0.90} \textbf{43.30{\scriptsize$\pm$0.71}} \\
    \bottomrule
  \end{tabular}%
  }
\end{table*}

\begin{table*}[t]
  \centering
  \caption{In-hospital mortality and 30-day readmission (mean $\pm$ std, 5 seeds). }
  \label{tab:clinical_tasks}
  \vspace{-2mm}
  \renewcommand{\arraystretch}{1.15}
  \resizebox{\textwidth}{!}{%
  \small
  \setlength{\tabcolsep}{10pt}
  \begin{tabular}{ll ccc ccc}

    \toprule
    \multicolumn{2}{c}{} & \multicolumn{3}{c}{\textbf{MIMIC-III (\%)}}
      & \multicolumn{3}{c}{\textbf{MIMIC-IV (\%)}} \\
    \cmidrule(lr){3-5}\cmidrule(lr){6-8}
    \textbf{Task} & \textbf{Model}
      & AUROC & AUPRC & F1
      & AUROC & AUPRC & F1 \\
    \cmidrule(l){1-8}

    \multirow{4}{*}{\makecell[l]{\textit{In-Hospital}\\\textit{Mortality}}}
      & GraphCare
      & 85.90{\scriptsize$\pm$0.25} & 47.78{\scriptsize$\pm$0.50} & 43.93{\scriptsize$\pm$2.23}
      & 87.16{\scriptsize$\pm$0.29} & 49.67{\scriptsize$\pm$0.75} & 44.57{\scriptsize$\pm$1.89} \\
      & RAM-EHR
      & \underline{86.57{\scriptsize$\pm$0.28}} & 48.61{\scriptsize$\pm$1.08} & 44.73{\scriptsize$\pm$1.57}
      & 87.30{\scriptsize$\pm$0.29} & \underline{50.22{\scriptsize$\pm$0.88}} & 44.00{\scriptsize$\pm$1.60} \\
      & KARE
      & 86.50{\scriptsize$\pm$0.32} & \underline{48.77{\scriptsize$\pm$0.62}} & \underline{45.35{\scriptsize$\pm$2.34}}
      & \underline{87.89{\scriptsize$\pm$0.22}} & 49.26{\scriptsize$\pm$0.57} & \underline{44.72{\scriptsize$\pm$1.87}} \\
      & \cellcolor[gray]{0.90} \textbf{\modelname{}}
      & \cellcolor[gray]{0.90} \textbf{89.16{\scriptsize$\pm$0.10}}
      & \cellcolor[gray]{0.90} \textbf{53.29{\scriptsize$\pm$0.53}}
      & \cellcolor[gray]{0.90} \textbf{47.21{\scriptsize$\pm$1.27}}
      & \cellcolor[gray]{0.90} \textbf{90.05{\scriptsize$\pm$0.19}}
      & \cellcolor[gray]{0.90} \textbf{53.80{\scriptsize$\pm$1.11}}
      & \cellcolor[gray]{0.90} \textbf{49.88{\scriptsize$\pm$2.09}} \\

    \cmidrule(l){1-8}
    \multirow{4}{*}{\makecell[l]{\textit{30-Day}\\\textit{Readmission}}}
      & GraphCare
      & 68.05{\scriptsize$\pm$1.65} & 40.26{\scriptsize$\pm$1.40} & 35.85{\scriptsize$\pm$1.10}
      & 68.95{\scriptsize$\pm$1.33} & 41.06{\scriptsize$\pm$1.04} & 36.23{\scriptsize$\pm$1.23} \\
      & RAM-EHR
      & \underline{68.32{\scriptsize$\pm$1.78}} & \underline{40.48{\scriptsize$\pm$1.32}} & 34.71{\scriptsize$\pm$2.22}
      & 68.74{\scriptsize$\pm$1.17} & 40.81{\scriptsize$\pm$0.93} & \underline{36.56{\scriptsize$\pm$1.56}} \\
      & KARE
      & 67.95{\scriptsize$\pm$1.88} & 40.19{\scriptsize$\pm$1.56} & \underline{36.31{\scriptsize$\pm$2.67}}
      & \underline{69.01{\scriptsize$\pm$1.75}} & \underline{41.06{\scriptsize$\pm$1.35}} & 35.28{\scriptsize$\pm$1.83} \\
      & \cellcolor[gray]{0.90} \textbf{\modelname{}}
      & \cellcolor[gray]{0.90} \textbf{70.91{\scriptsize$\pm$1.13}}
      & \cellcolor[gray]{0.90} \textbf{43.01{\scriptsize$\pm$1.08}}
      & \cellcolor[gray]{0.90} \textbf{40.12{\scriptsize$\pm$0.75}}
      & \cellcolor[gray]{0.90} \textbf{73.57{\scriptsize$\pm$0.96}}
      & \cellcolor[gray]{0.90} \textbf{45.29{\scriptsize$\pm$0.95}}
      & \cellcolor[gray]{0.90} \textbf{42.09{\scriptsize$\pm$2.05}} \\
    \bottomrule
  \end{tabular}%
  }
  \vspace{-5pt}
\end{table*}

\subsection{Setup}
\label{sec:setup}

\paragraph{Datasets.} We evaluate on MIMIC-III~\cite{johnson2016mimic} and MIMIC-IV~\cite{johnson2023mimic} for next-visit diagnosis prediction (CCS granularity), in-hospital mortality, and 30-day readmission. Dataset statistics are in Appendix~\ref{app:A1_dataset}.

\paragraph{Baselines.} We compare nine baselines across four categories. Sequence models include Transformer~\cite{vaswani2017attention} and RETAIN~\cite{choi2016retain}. Ontology encoders include G-BERT~\cite{shang2019pre} and HAP~\cite{zhang2020hierarchical}. Structure learners include GRAM~\cite{choi2017gram} and SeqCare~\cite{xu2023seqcare}. KG-augmented systems include GraphCare~\cite{jiang2024graphcare}, RAM-EHR~\cite{xu2024ram}, and KARE~\cite{ICLR2025_cb5a3b45}. All methods share the data splits and ICD inputs, with KG-augmented methods also sharing PrimeKG and the ICD-to-KG linker (Appendix~\ref{app:A1_dataset}). We additionally evaluate GPT-4o~\cite{hurst2024gpt} by providing ICD descriptions of $V_{1:t}$ and requesting a score in $[0,1]$ for each CCS label. Its +\,PrimeKG variant adds each code's 1-hop KG neighbors to the prompt.

\paragraph{Metrics.} We report macro-AUPRC (primary, abbreviated as AUPRC), Micro-F1, and Acc@20 for diagnosis prediction, and AUROC, AUPRC, and F1 for mortality and readmission. For GPT-4o, we perform five generation runs at temperature 0.2, each producing one label-score vector per prediction instance. We compute metrics separately for each run and report their mean and standard deviation. Micro-F1 uses a threshold selected on validation. Hyperparameter settings are in Appendix~\ref{app:implementation}. Differences are computed before rounding.

\subsection{Main Results}
\label{sec:main_results}

\paragraph{Diagnosis prediction.}
\modelname{} achieves the highest mean scores on both datasets across all three metrics (Table~\ref{tab:main_results}), with AUPRC gains of +1.81 and +2.08 over KARE on MIMIC-III and MIMIC-IV.
\modelname{} also achieves the highest AUPRC in both transfer directions (Table~\ref{tab:main_results}).
KG-augmented methods remain competitive, although G-BERT ranks above GraphCare and RAM-EHR in AUPRC on MIMIC-IV.
Matched controls in \S\ref{sec:ablation} examine the contribution of visit-level augmentation decisions.

\paragraph{LLM and KG generality.}
GPT-4o scores below all non-LLM baselines, including RETAIN.
The gap may reflect the difficulty of producing calibrated predictions over 283 CCS categories from next-token generation.
Adding PrimeKG context improves performance, but a gap remains relative to graph-based prediction.
\modelname{} uses the LLM for offline distillation, where clinical descriptions complement the graph encoder.
The UMLS variant also achieves higher AUPRC and Acc@20 than all baselines on both datasets, showing that the framework works with either knowledge source.
PrimeKG performs better, consistent with its denser pathophysiological links.

\paragraph{Generalization.}
All methods lose AUPRC when transferred between datasets, but \modelname{} loses less in relative terms in both directions.
These results suggest that the knowledge-augmented representations retain useful information across the two datasets.
On MIMIC-III, \modelname{}'s transfer result even exceeds KARE's in-domain result (Table~\ref{tab:main_results}).

\paragraph{Mortality and readmission.}
\modelname{} improves over the strongest baselines on both binary tasks, with the largest AUROC margin on MIMIC-IV readmission (Table~\ref{tab:clinical_tasks}).
Interactions among chronic conditions may help explain this readmission gain, whereas mortality depends more on acute severity already captured by the base encoder.

\subsection{Ablation}
\label{sec:ablation}
\begin{figure}[t]
  \centering
  \includegraphics[width=0.95\linewidth]{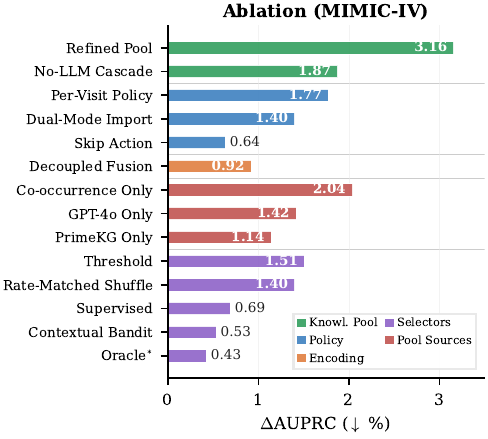}
  \caption{Ablation on MIMIC-IV, reported as AUPRC drop from the full model. $^{*}$Label-informed myopic diagnostic (see \S\ref{sec:ablation}). Additional component and selector results are in Appendix~\ref{app:extended_ablation}.}
  \label{fig:ablation}
\end{figure}

\paragraph{Component ablation.}
The combined pool outperforms all three alternatives in Figure~\ref{fig:ablation}, with the largest AUPRC drop under Co-occurrence Only and smaller drops under GPT-4o Only and PrimeKG Only.
Replacing LLM-distilled cascades with CCS-hierarchy expansions also reduces accuracy, showing the value of the distilled content within the pool.
The per-visit policy and the two import modes contribute to performance, while freezing the encoder during Stage~2 has a smaller effect.
The rate-matched shuffle lowers AUPRC on both datasets despite keeping the checkpoint, updates, budget and action rates fixed.
This comparison highlights the value of choosing which visits receive each action.

\paragraph{Selector comparison.} The learned policy achieves the highest AUPRC among deployable selectors with the same pool and encoder (Figure~\ref{fig:ablation} and Table~\ref{tab:selectors}). The contextual bandit comes closest at similar latency. The threshold heuristic preserves template rankings, while the supervised selector requires exhaustive action labels. The greedy oracle uses true test labels to minimize per-visit cross-entropy. Macro-AUPRC measures rankings across visits and averages over labels, so minimizing per-visit loss need not maximize it.

\paragraph{Pool sources and validation.} 
The full \modelname{} pool combines LLM-generated content with structural support from PrimeKG or CCS for every retained Hard-Import relation (Appendix~\ref{app:grounding_details}). An audit of 240 templates finds high definition and relation accuracy, with more major errors among rare diagnoses (Appendix~\ref{app:audit}). Rebuilding the pool with Qwen3-32B retains most grounded links and yields similar AUPRC (Appendix~\ref{app:generator}).

\subsection{Analysis}
\label{sec:analysis}

\begin{figure}[t]
  \centering
  \includegraphics[width=\linewidth]{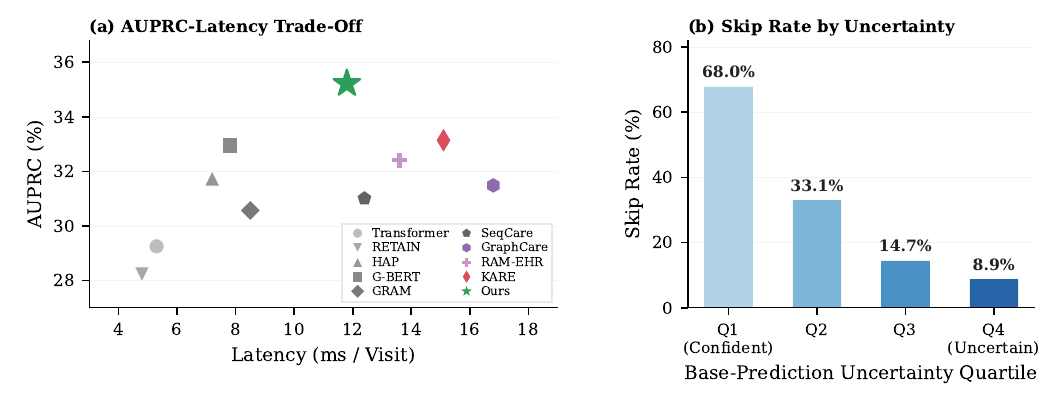}
  \caption{AUPRC vs.\ per-visit latency on MIMIC-IV (left) and per-visit skip rate by base-prediction uncertainty quartile (right).}
  \label{fig:analysis}
  \vspace{-5pt}
\end{figure}

\begin{figure*}[t]
  \centering
  \includegraphics[width=\linewidth]{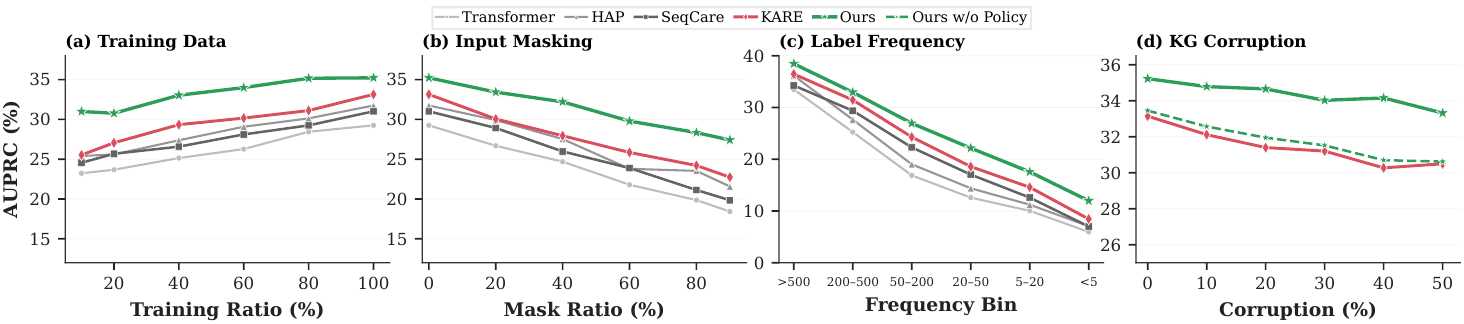}
  \caption{Performance on MIMIC-IV under reduced training data (a), diagnosis masking (b), label-frequency stratification (c), and KG corruption (d). Baselines that do not consume the corrupted graph are omitted from (d).}
  \label{fig:robustness}
\end{figure*}

\paragraph{Latency.}
\modelname{} combines higher AUPRC with lower latency than GraphCare and KARE (Figure~\ref{fig:analysis}a).
Skipped visits avoid graph assembly, limiting the added cost of policy evaluation.

\paragraph{Skip behavior.}
Skip is most frequent in the most confident quartile and becomes less common as uncertainty rises (Figure~\ref{fig:analysis}b).
The variant trained without Skip has lower AUPRC and higher latency (Table~\ref{tab:selectors}).
Sensitivity and calibration results are in Appendices~\ref{app:sensitivity} and~\ref{app:calibration}.

\paragraph{Retrieval overrides.} The policy departs from code-only Top-1 retrieval on roughly half of visits and changes the template on about a third of non-Skip visits. Forcing Top-1 selection lowers AUPRC on both datasets (Table~\ref{tab:overrides}). These results show broad use of history-aware routing.

\subsection{Robustness}
\label{sec:robustness}

\paragraph{Training data and input completeness.} 
\modelname{}'s margin over KARE widens as less training data is available (Figure~\ref{fig:robustness}a). 
For input masking, we independently remove each diagnosis code from the observed history $V_{1:t}$ with the specified probability at test time. If all codes in a visit are masked, we retain one uniformly sampled original code. Target labels and trained parameters remain fixed. Figure~\ref{fig:robustness}b shows that \modelname{} retains higher AUPRC under diagnosis masking, with the largest margin over KARE at the highest masking ratio.

\paragraph{Label frequency and knowledge quality.} 
For KG corruption, we randomly select the specified fraction of external knowledge edges and replace one endpoint with a uniformly sampled node from the same graph. We perturb the knowledge subgraphs supplied to each model at test time, including \modelname{}'s template subgraphs before import, while keeping the CCS hierarchy, patient inputs, target labels, and trained parameters fixed.
The gain over KARE generally increases toward the tail (Figure~\ref{fig:robustness}c). Under KG corruption, \modelname{} loses less AUPRC than its variant without the policy (panel~d). Per-stratum results are in Appendix~\ref{app:label_freq}.

\paragraph{Richer structured inputs.}
With prior medication and procedure codes added under the same cutoff, \modelname{} remains more accurate and faster than KARE (Table~\ref{tab:richer}).
Hard Import becomes less frequent, suggesting that richer inputs reduce the need for structural grafting.
The evaluation covers structured codes, and clinical notes remain future work.

\subsection{Case Study}
\label{sec:case_study}

\begin{figure}[t]
  \centering
  \includegraphics[width=\linewidth]{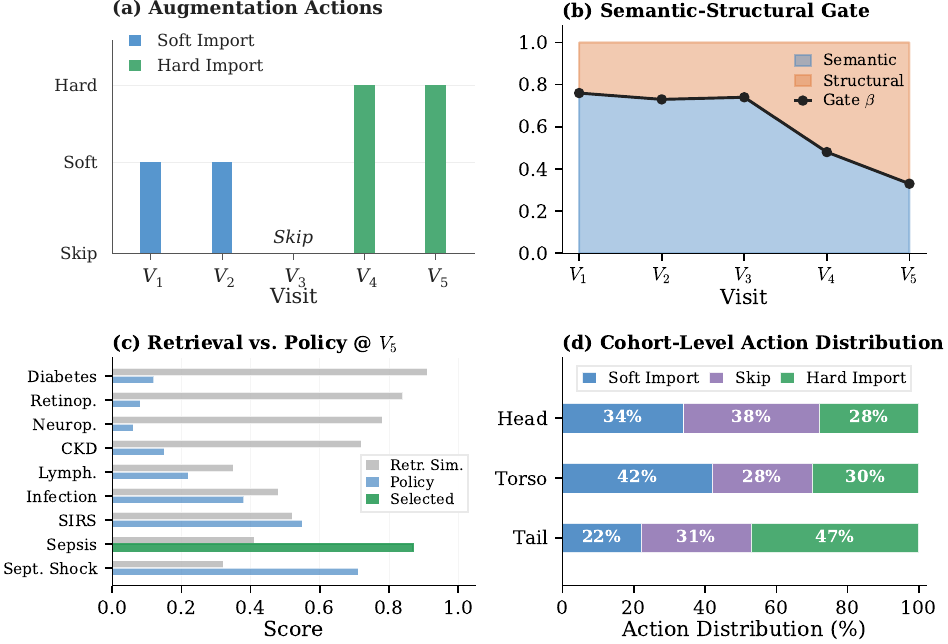}
  \caption{Augmentation actions (a), fusion gate $\beta$ (b), retrieval and policy scores (c), and action distributions for head ($>200$), torso ($20$--$200$), and tail ($<20$ occurrences) diagnoses (d).}
  \label{fig:case_study}
\end{figure}

\begin{figure}[t]
    \centering
    \includegraphics[width=\linewidth]{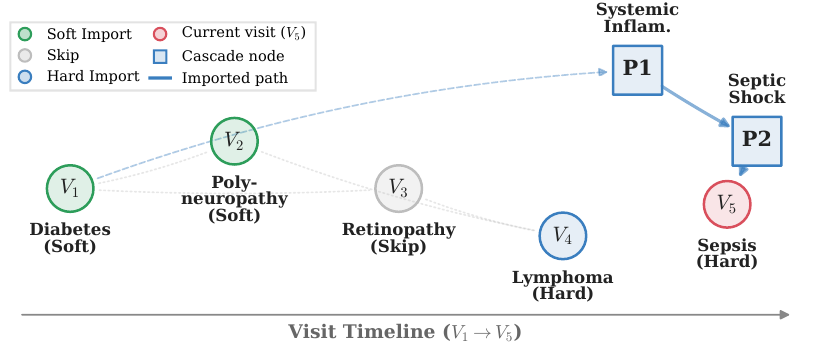}
    \caption{Hard Import at $V_5$. Solid blue arrows show imported paths within the current visit graph, while the dashed blue arrow represents historical context from across-visit attention.}
    \label{fig:topology_viz}
\end{figure}

Figure~\ref{fig:case_study} traces how the policy adapts as the clinical state evolves across five observed visits. 
Next-visit prediction uses the four transitions ending at $V_2$--$V_5$. We also examine Hard Import at $V_5$, the last observed visit. Soft Import is applied at $V_1$-$V_2$ (routine chronic visits), $V_3$ is skipped where the base prediction is already confident ($u_3{=}0.14$), and Hard Import activates at $V_4$-$V_5$ as the state shifts acutely. 
The fusion gate in panel~(b) shifts from semantic-dominant ($\beta{=}0.76$ at $V_1$) to structure-dominant ($\beta{=}0.33$ at $V_5$), with the sharpest transition at the chronic-to-acute boundary.
In panel~(c), code-level retrieval ranks chronic templates highest, while the policy selects the Sepsis template at $V_5$. Figure~\ref{fig:topology_viz} shows cascade nodes enriching the current visit, with earlier chronic context supplied through across-visit attention. In panel~(d), Skip is most frequent in the head group and Hard Import in the tail group (Appendix~\ref{sec:appendix_c_policy}).

\section{Related Work}
\label{sec:related_work}
 
\paragraph{EHR prediction.}
Longitudinal EHR prediction has evolved from recurrent models~\cite{choi2016retain} to Transformer-style architectures~\cite{luo2020hitanet,rasmy2021med}, but purely data-driven models degrade under sparsity, motivating external knowledge as complementary priors.
 
\paragraph{KG augmentation.}
Topology-oriented methods expand visit graphs with external
structure.
GRAM~\cite{choi2017gram} and G-BERT~\cite{shang2019pre} use predefined ontology hierarchies, while SeqCare~\cite{xu2023seqcare} incorporates external medical knowledge and refines the graph structure during sequential training.
GraphCare~\cite{jiang2024graphcare} builds patient-specific KGs from LLM-generated and biomedical knowledge.
KARE~\cite{ICLR2025_cb5a3b45} and RAM-EHR~\cite{xu2024ram} use retrieved knowledge to support EHR prediction.
\modelname{} focuses on jointly selecting a template, its import mode, or Skip at each visit under an augmentation budget.
 
\paragraph{Selective augmentation.}
RL-based KG reasoning~\cite{das2018go,xiong2017deeppath} learns paths over a fixed, pre-existing graph rather than deciding how to augment a patient-specific one.
Learnable prompts~\cite{liao2025learnable} inject soft tokens into the feature space without modifying topology or offering abstention.
\modelname{} coordinates topology and semantic augmentation with abstention under a per-visit budget, selecting quality-filtered templates as the patient's state evolves.

\section{Conclusion}
\label{sec:conclusion}

\modelname{} treats knowledge augmentation in longitudinal EHR prediction as a budgeted, per-visit decision, selecting Soft Import, Hard Import, or Skip from a quality-filtered knowledge pool based on each patient's evolving clinical state.
Experiments on MIMIC-III and MIMIC-IV show consistent gains across diagnosis prediction, mortality, and readmission, with the advantage growing under sparse supervision and persisting under cross-dataset transfer and KG replacement (UMLS).
Future work could extend the framework to multi-modal EHR inputs and uncertainty-aware retrieval.

\section*{Limitations}
 
\modelname{} relies on external KGs (PrimeKG, UMLS), which may be incomplete for some clinical domains, and on LLM-generated cascades that can produce plausible but incorrect complications for rare diseases, where the audit finds a higher major-error rate (2.5\% vs.\ 0.8\%, Appendix~\ref{app:audit}).
Grounding establishes structural support rather than clinical correctness, and because GPT-4o's pretraining corpus is undisclosed, indirect overlap with MIMIC-derived literature cannot be ruled out.
The skip mechanism assumes base-encoder confidence is a reliable proxy for augmentation need, which fails when the encoder is confidently wrong (Appendix~\ref{sec:appendix_c_qualitative}).
Finally, our evaluation covers structured codes and does not incorporate clinical notes.
Evaluation on disjoint patient cohorts remains future work.
 
\section*{Ethics Statement}
 
All experiments use MIMIC-III and MIMIC-IV under the PhysioNet Credentialed Health Data License.
The datasets contain de-identified patient records, and no re-identification was attempted.
\modelname{} is a research prototype not intended for clinical deployment without extensive validation, regulatory review, and integration with clinical oversight.
Model performance may vary across patient subgroups defined by age, sex, race, or insurance status, and no fairness audit was conducted in this work.
LLM-generated knowledge templates are filtered by ontology grounding and assessed by a blinded clinical audit (Appendix~\ref{app:audit}), but residual errors may propagate into predictions for underrepresented conditions.

\section*{Acknowledgments}
This work was supported by the National Science Foundation under Grant No.~OIA-2531881. We also thank the anonymous reviewers for their valuable suggestions and feedback.

\bibliography{main}

\appendix
\clearpage

\setcounter{topnumber}{4}
\setcounter{bottomnumber}{4}
\setcounter{totalnumber}{8}
\renewcommand{\topfraction}{0.95}
\renewcommand{\bottomfraction}{0.95}
\renewcommand{\textfraction}{0.03}
\renewcommand{\floatpagefraction}{0.85}

\newenvironment{apptable}{
  \small
  \setlength{\tabcolsep}{4pt}
  \renewcommand{\arraystretch}{1.02}
}{}

\newcommand{\apptocsec}[2]{\par\medskip\noindent\textbf{\ref{#1}\enspace #2}\hfill\textbf{\pageref{#1}}\par\smallskip}
\newcommand{\apptocsub}[2]{\noindent\hspace*{1.5em}\ref{#1}\enspace #2\ \dotfill\ \pageref{#1}\par}

\section*{Appendix Contents}
\begingroup\small
\apptocsec{sec:appendix_a}{Reproducibility Details}
\apptocsub{app:A1_dataset}{Dataset and Preprocessing}
\apptocsub{app:implementation}{Implementation and Hyperparameters}
\apptocsec{sec:appendix_b}{Knowledge Pool Construction}
\apptocsub{app:B1_topology}{CCS Hierarchy and Hard Import}
\apptocsub{app:B3_distill}{LLM Distillation}
\apptocsub{app:grounding_details}{Grounding and Filtering}
\apptocsub{app:B2_action_schema}{Template Construction}
\apptocsub{app:pool_diagnostics}{Pool Diagnostics}
\apptocsub{app:audit}{Clinical Audit}
\apptocsub{app:generator}{Generator Dependence}
\apptocsec{app:sensitivity}{Sensitivity}
\apptocsec{app:extended_ablation}{Extended Ablation Results}
\apptocsec{app:additional_results}{Additional Results}
\apptocsub{app:label_freq}{Label-Frequency Stratification}
\apptocsub{app:calibration}{Calibration}
\apptocsec{sec:appendix_c}{Mechanism Analysis}
\apptocsub{sec:appendix_c_policy}{Policy Behavior}
\apptocsub{sec:appendix_c_paired}{Reward Comparison}
\apptocsub{sec:appendix_c_qualitative}{Qualitative Analysis}
\endgroup

\section{Reproducibility Details}
\label{sec:appendix_a}

\subsection{Dataset and Preprocessing}
\label{app:A1_dataset}

\begin{table}[htbp]
  \centering
  \begin{apptable}
  \caption{Dataset statistics for the retained cohort after preprocessing. CCS labels are deduplicated within each visit.}
  \label{tab:dataset_stats}
  \begin{tabular}{lrr}
    \toprule
    \textbf{Metric} & \textbf{MIMIC-III} & \textbf{MIMIC-IV} \\
    \midrule
    \# Patients (eligible)       & 7{,}023  & 94{,}402  \\
    \# Visits (records)          & 18{,}457 & 250{,}949 \\
    \# Transitions (samples)     & 11{,}434 & 156{,}547 \\
    Avg.\ visits / patient       & 2.63     & 2.66      \\
    Visits/pat.\ (P25/P50/P75)   & 2/2/3    & 2/2/3     \\
    \# Unique ICD codes $|\mathcal{C}_{dx}|$ & 4{,}137 & 8{,}507 \\
    Avg.\ diagnoses / visit      & 13.08    & 21.62     \\
    Avg.\ CCS labels / visit     & 4.91     & 6.27      \\
    Label space $|\mathcal{A}|$  & \multicolumn{2}{c}{\(283\)} \\
    \bottomrule
  \end{tabular}
  \end{apptable}
\end{table}

\paragraph{Cohort construction.}
We evaluate on MIMIC-III (ICD-9) and MIMIC-IV (ICD-10), with statistics in Table~\ref{tab:dataset_stats} (P25, P50, P75 denote percentiles).
MIMIC-IV spans the ICD-9 to ICD-10 transition, so we retain only admissions coded in ICD-10.
Diagnosis codes in MIMIC are recorded at the hospital-admission level.
Each visit is represented by a set of diagnosis codes with duplicates removed, and we retain patients with at least two visits.
For a patient trajectory with $T_i$ visits, we construct $(T_i-1)$ samples for next-visit prediction.

\paragraph{Splits.}
All methods share the same fixed splits based on deidentified discharge times.
Admissions discharged in the earliest 70\% of the deidentified time range form the training set, the next 10\% form validation, and the final 20\% form the test set.
Split membership is assigned at the admission level.
Dates are shifted separately for each patient, so this partition does not constitute a calendar-time holdout.
Split files are fixed once and shared by all methods, and model selection is performed on validation only.

\paragraph{Prediction targets.} ICD-9 and ICD-10 diagnosis codes are mapped with AHRQ CCS~\cite{hcup2017ccs} and beta CCS for ICD-10-CM~\cite{hcup2019ccs10}, respectively. Targets are distinct CCS categories at the next visit in the shared label space $\mathcal{A}$ (Table~\ref{tab:dataset_stats}). Only diagnoses are used as inputs and for visit-graph construction, except in the richer-input experiment in \S\ref{sec:robustness} (Table~\ref{tab:richer}). Online message passing uses an untyped, symmetrized adjacency.

\paragraph{Additional tasks.} For mortality prediction, diagnosis codes from $V_{1:t}$ predict in-hospital death during $V_{t+1}$. Patient-specific graphs and retrieval queries use the same observed history. For 30-day readmission, the model predicts whether another admission occurs within 30 days after discharge from $V_t$. Both tasks share the encoder and knowledge pool, with a sigmoid output trained using binary cross-entropy.

\paragraph{Cross-dataset transfer.}
For the transfer rows in Table~\ref{tab:main_results}, each method is trained on one dataset and evaluated on the other without retraining. 
The prediction targets use the shared CCS label space $\mathcal{A}$.
GraphCare and RAM-EHR's local EHR branch represent diagnosis codes using a frozen ClinicalBERT encoder over canonical descriptions followed by a projection learned on the source dataset. RAM-EHR's text branch and KARE receive canonical diagnosis descriptions. Each method uses the same input representation during source training and target evaluation. Learned parameters, knowledge resources, and the CCS output mapping remain fixed during target evaluation.

For \modelname{}, source and target ICD codes are encoded from their canonical descriptions using the same frozen ClinicalBERT encoder and the projection learned on the source dataset (\S\ref{sec:retrieval}). 
Target visit graphs follow the same ICD-to-CCS and PrimeKG linking procedure, with shared graph nodes merged by canonical identifiers. 
The source-trained model, template utilities, and knowledge pool remain fixed during target evaluation.
Transfer from MIMIC-IV to MIMIC-III uses the MIMIC-IV pool, while the reverse direction uses the MIMIC-III pool. 
Their sizes are reported in Table~\ref{tab:prompt_pool_quality}.

\subsection{Implementation and Hyperparameters}
\label{app:implementation}

\begin{table}[htbp]
  \centering
  \begin{apptable}
  \caption{Hyperparameter settings.}
  \label{tab:hyperparams}
  \begin{tabular}{llc}
    \toprule
    \textbf{Module} & \textbf{Parameter} & \textbf{Value} \\
    \midrule
    \multirow{4}{*}{General}
      & Embedding dim $d$ & 256 \\
      & Batch size & 32 \\
      & Learning rate & \(10^{-4}\) \\
      & Weight decay & \(10^{-5}\) \\
    \addlinespace
    \multirow{3}{*}{Backbone}
      & GNN layers $L$ & 2 \\
      & Attention heads & 4 \\
      & Dropout & 0.3 \\
    \addlinespace
    \multirow{3}{*}{Training}
      & Stage 1 epochs $E_{\mathrm{pre}}$ & 30 \\
      & Stage 2 iterations $I_{\mathrm{rl}}$ & 50 \\
      & Exposure rate (Stage 1) & 0.3 \\
    \addlinespace
    \multirow{5}{*}{Policy}
      & Discount factor $\gamma$ & 0.95 \\
      & Retrieval size $K$ & 20 \\
      & Reward weight $\lambda_1$ & 1.0 \\
      & Hard Import penalty $\lambda_2$ & 0.1 \\
      & Baseline decay & 0.99 \\
    \addlinespace
    \multirow{2}{*}{Policy update}
      & Policy learning rate & \(10^{-5}\) \\
      & Max grad norm & 0.5 \\
    \bottomrule
  \end{tabular}
  \end{apptable}
\end{table}

\paragraph{Hardware and training time.} Experiments use Adam on 2$\times$NVIDIA L40S GPUs (48\,GB each). Stage~1 converges in about 3 hours and Stage~2 in 8 hours, totaling ${\sim}$11 hours. Hyperparameters (Table~\ref{tab:hyperparams}) are selected on validation.

\paragraph{Released defaults.}
The released implementation sets the retrieval weight $\alpha=0.3$, pool cut $\tau=0.16$, Soft Import scale $\xi=0.5$, template utility decay $\gamma_r=0.95$, and CCS ancestor depth $h=2$.

\paragraph{Policy optimization.} A lightweight MLP on $s_t$ defines a categorical policy over $2K{+}1$ actions ($K$ templates $\times$ 2 modes plus Skip). We use REINFORCE~\cite{zhang2021sample} with a running-mean baseline (decay 0.99) to reduce variance. With short trajectories (median 2 visits, 75th percentile 3) and a lightweight policy, critic-based methods are unnecessary (Algorithm~\ref{alg:training}).

\paragraph{Ablation variants.} 
No-LLM Cascade uses CCS expansions up to two ancestor levels per code. Frozen Encoder updates only the policy during Stage~2. The uncertainty heuristic assigns Skip, Soft Import, and Hard Import for $u_t{<}0.3$, $0.3{\leq}u_t{<}0.6$, and $u_t{\geq}0.6$, respectively, using the top-ranked template from Eq.~\ref{Eq:score}. The supervised selector trains an MLP on per-visit minimum-loss action labels obtained by evaluating all $2K{+}1$ actions during training. The greedy myopic oracle performs this selection using test labels.

The \textit{Per-Visit Policy} ablation reuses the first decision's action throughout the trajectory. \textit{Dual-Mode Import} fixes one validation-selected import mode for all visits, retaining per-visit template selection and Skip.

The pool-source variants retain the encoder, retrieval size $K$, and import budget. \textit{Co-occurrence Only} builds cascades from diagnosis co-occurrence within training visits. \textit{PrimeKG Only} uses PrimeKG neighborhoods without LLM distillation. Both use canonical descriptions as template text. \textit{GPT-4o Only} uses LLM-generated definitions and cascades with entity normalization but no external relation verification.

\textit{Contextual Bandit} retains the full policy's state, retrieval, and action space, optimizing only the immediate paired reward $r_t$. \textit{Gate-only fusion} encodes raw and augmented visit graphs with a shared GAT and across-visit attention, then combines their visit representations with a learned sigmoid gate. \textit{Forced Top-1} substitutes the global pool's code-only Top-1 template for each non-Skip action, preserving the policy's Skip decisions and import modes.

\paragraph{Inference latency.}
Table~\ref{tab:latency_breakdown} shows where inference time is spent.
We measure end-to-end inference latency at batch size 1, including retrieval, online graph or context construction, and prediction. Offline knowledge-pool construction is excluded.
KARE predicts the 283 CCS label scores with a sigmoid head over the final input-token representation, retaining reasoning supervision during training. We measure label prediction without reasoning-chain generation.
\modelname{}'s skip-eligible visits avoid graph assembly entirely (0.0\,ms), reducing the average per-visit cost below both GraphCare and KARE despite the additional policy forward pass.

\begin{table}[htbp]
  \centering
  \begin{apptable}
  \setlength{\tabcolsep}{3.5pt}
  \caption{Inference latency per visit on MIMIC-IV (ms).}
  \label{tab:latency_breakdown}
  \begin{tabular}{lcccc@{\hskip 6pt}c}
    \toprule
    \textbf{Method} & \textbf{Retr.} & \textbf{Policy} & \textbf{Graph} & \textbf{Enc.} & \textbf{Total} \\
    \midrule
    GraphCare    & 4.2 & --  & 6.8 & 5.8 & 16.8 \\
    KARE & -- & -- & -- & -- & 15.1 \\
    \modelname{} & 2.6 & 1.4 & 2.3 & 5.5 & 11.8 \\
    \quad w/ Skip & 2.6 & 1.4 & 0.0 & 5.5 & 9.5 \\
    \bottomrule
  \end{tabular}
  \end{apptable}
\end{table}

\begin{algorithm}[htbp]
\small
\caption{Two-stage training curriculum.}
\label{alg:training}
\begin{algorithmic}[1]
\REQUIRE Trajectories \(\mathcal{D}\), global knowledge pool \(\mathcal{P}_{\mathrm{global}}\), with \(T_\tau\) visits per trajectory for next-visit diagnosis prediction
\ENSURE Policy \(\pi_\theta\), encoder \(f_\psi\)

\STATE \textbf{Stage 1. Encoder warm-up}
\FOR{\(e=1\) to \(E_{\mathrm{pre}}\)}
  \FOR{batch \(\mathcal{B}\subset\mathcal{D}\)}
    \STATE Construct visit graphs \(\{G_t\}\) for \(\mathcal{B}\)
    \STATE Set \(G'_t \leftarrow G_t\) \COMMENT{no import by default}
    \STATE Sample exposure flag \(u \sim \mathrm{Bernoulli}(0.3)\)
    \IF{\(u=1\)}
      \STATE Retrieve candidates \(\mathcal{P}^{(t)}_{\mathrm{sub}}\) (Top-\(K\) by Eq.~\ref{Eq:score})
      \STATE Sample \(\tilde a_t \sim \mathrm{Uniform}(\mathcal{P}^{(t)}_{\mathrm{sub}}\times\{0,1\})\)
      \STATE Apply \(\tilde a_t\) to obtain \(G'_t\) \COMMENT{stochastic Soft/Hard}
    \ENDIF
    \STATE Update \(\psi \leftarrow \arg\min_\psi \mathcal{L}_{\mathrm{task}}(f_\psi;G'_t)\)
  \ENDFOR
\ENDFOR

\STATE \textbf{Stage 2. Policy learning (REINFORCE)}
\FOR{\(i=1\) to \(I_{\mathrm{rl}}\)}
  \STATE Clear rollout buffer \(\mathcal{M}\)
  \FOR{trajectory \(\tau \sim \mathrm{Sample}(\mathcal{D})\)}
    \FOR{\(t=1\) to \(T_\tau-1\)}
      \STATE Observe \(s_t \gets \mathrm{GRU}(v_{1:t-1}) \oplus v_t\)
      \STATE Compute \(u_t\) with the frozen Stage~1 encoder and set \(s_t \gets s_t \oplus u_t\)
      \STATE Retrieve \(\mathcal{P}^{(t)}_{\mathrm{sub}}\) using Eq.~\ref{Eq:score}
      \STATE Form \(s_t^\pi \gets s_t \oplus (\bar{r}_k)_{\mathbf{p}_k \in \mathcal{P}^{(t)}_{\mathrm{sub}}}\)
      \STATE Sample \(a_t \sim \pi_\theta(\cdot \mid s_t^\pi)\)
      \STATE Execute \(a_t\) to obtain \(G'_t\)
      \STATE Compute the paired reward \(r_t\) with dropout disabled
      \STATE Store \((a_t,r_t,\log\pi_\theta(a_t\mid s_t^\pi))\) in \(\mathcal{M}\)
      \IF{\(a_t \neq \texttt{Skip}\)}
        \STATE Update the selected template's utility \(\bar{r}_k \gets \gamma_r\bar{r}_k+(1-\gamma_r)r_t\)
      \ENDIF
    \ENDFOR
  \ENDFOR
  \STATE Compute baseline-subtracted returns \(\hat{R}_t = R_t - \bar{R}\)
  \STATE Update \(\theta\) via \(\nabla_\theta \mathcal{J}^{\mathrm{RL}} = \mathbb{E}[\hat{R}_t \nabla_\theta \log \pi_\theta(a_t \mid s_t^\pi)]\)
  \STATE Update \(\psi\) to minimize \(\mathcal{L}_{\mathrm{task}} - \eta \mathcal{J}^{\mathrm{RL}}\)
\ENDFOR
\end{algorithmic}
\end{algorithm}

\paragraph{Complexity.}
Retrieval supplies $K$ candidates, but at most one template is imported per visit.
Let $\bar{n}$ and $\bar{e}$ denote the average numbers of nodes and edges in the selected template.
Using the average template sizes $\bar{n}$ and $\bar{e}$, the encoder cost is $O(L((|V|+\bar{n})d^2+(|E|+\bar{e})d))$ for $L$ GAT layers.
Template scoring costs $\mathcal{O}(|V_t||\mathcal{P}_{\mathrm{global}}|d)$ per visit.

\section{Knowledge Pool Construction}
\label{sec:appendix_b}

All steps in this section are executed offline from external resources and distilled artifacts only, with no access to patient trajectories, outcome labels, or split-dependent statistics.

\subsection{CCS Hierarchy and Hard Import}
\label{app:B1_topology}

\begin{figure}[htbp]
    \centering
    \includegraphics[width=\linewidth]{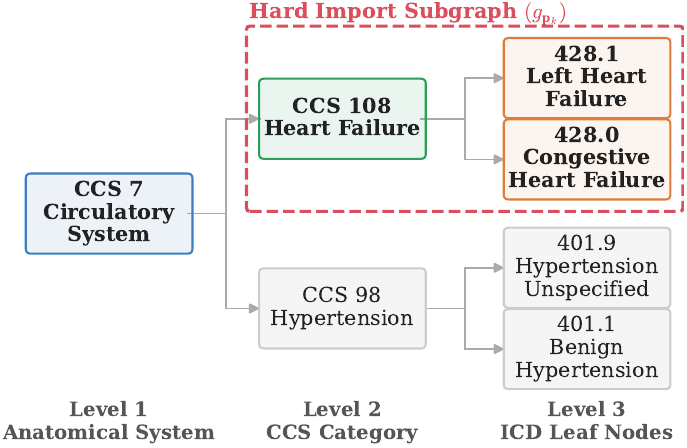}
    \caption{CCS hierarchy with a Hard Import graft example.}
    \label{fig:ccs_hierarchy}
\end{figure}

ICD codes form the leaf nodes of the diagnosis-side ontology graph (Figure~\ref{fig:ccs_hierarchy}), connected to CCS categories via ICD-to-CCS links.
CCS categories are connected by is-a edges.
Online message passing treats all links as an untyped, symmetrized adjacency, so shared ancestors act as short bridges across related diagnoses.
Hard Import exploits this structure by grafting a small subgraph containing shared CCS ancestors, creating short message-passing paths between otherwise weakly connected codes with minimal added edges.

\subsection{LLM Distillation}
\label{app:B3_distill}

\begin{figure}[htbp]
  \centering
  \setlength{\fboxrule}{0pt}
  \colorbox{blue!5}{%
    \parbox{0.96\columnwidth}{%
      \vspace{4pt}
      {\small\sffamily\bfseries\color{blue!70!black} System}\\[2pt]
      {\small You are an expert clinical pathologist. Given the diagnosis below, provide a structured knowledge summary to assist in sequential predictive modeling.}\\[6pt]
      {\small\sffamily\bfseries\color{blue!70!black} Input}\\[2pt]
      {\small \textbf{Diagnosis} \texttt{<ICD description>}}\\
      {\small \textbf{PrimeKG neighborhood density}\\ \texttt{<sparse\,|\,moderate\,|\,dense>}}\\[6pt]
      {\small\sffamily\bfseries\color{blue!70!black} Output Constraints}\\[2pt]
      {\small 1.\ \textbf{Definition} (for Soft Import). One sentence focusing on pathology.}\\
      {\small 2.\ \textbf{Clinical Cascade} (for Hard Import). \{1--5\} downstream complications}\\
      {\small \hspace{12pt}(sparse neighborhoods $\rightarrow$ up to 5, dense $\rightarrow$ as few as 1).}\\
      \vspace{3pt}
    }%
  }
  \caption{Offline distillation prompt with adaptive cascade length.}
  \label{fig:prompt_template}
\end{figure}

We use GPT-4o~\cite{hurst2024gpt} to distill templates from ICD diagnosis codes and their corresponding CCS categories using the prompt in Figure~\ref{fig:prompt_template}. Decoding uses temperature $0.2$, Top-$p=0.9$, and a maximum of 256 tokens. ICD descriptions come from dataset-provided dictionaries, with the mapped CCS category name used when a description is missing.

\subsection{Grounding and Filtering}
\label{app:grounding_details}

Each distilled mention is grounded to a canonical biomedical identifier by exact matching against ICD, CCS, and PrimeKG~\cite{chandak2023building} node vocabularies.
If exact matching fails, we compute cosine similarity between $\ell_2$-normalized name embeddings from ClinicalBERT~\cite{alsentzer2019publicly} (cached offline) and accept the Top-1 candidate only if similarity exceeds $\tau_{\text{map}}=0.90$.
Otherwise, the mention is discarded.

A distilled triple is retained only if both endpoints are groundable and the relation is supported by external ontology evidence, either via a direct PrimeKG edge or an ancestor/descendant relation within two CCS hierarchy levels.
All thresholds are fixed once and applied consistently across datasets.

\subsection{Template Construction}
\label{app:B2_action_schema}

Each grounded cascade is materialized into a template subgraph $g_{\mathbf{p}_k}=(V_{\mathbf{p}_k},E_{\mathbf{p}_k})$ that may include ICD leaves, CCS category nodes, and cascade entities.
We de-duplicate nodes by canonical identifier, drop self-loops, and merge duplicate edges.
Online aggregation collapses multi-relational links into one untyped edge per node pair.

The grounded outputs are then compressed by clustering.
Each concept text (definition concatenated with cascade) is embedded with ClinicalBERT and grouped via agglomerative clustering under cosine distance with cut threshold $\tau$.
Each template vector $\mathbf{p}_k$ is set to the $\ell_2$-normalized cluster centroid projected into the model embedding space.
The representative subgraph is the cluster medoid (closest to the centroid).
Three representative examples are shown in Table~\ref{tab:prompt_mapping}.

\begin{table}[htbp]
  \centering
  \begin{apptable}
  \caption{Knowledge template examples. Each template pairs a
  definition (for Soft Import) with a clinical cascade (for Hard
  Import).}
  \label{tab:prompt_mapping}
  \begin{tabularx}{\columnwidth}{@{} l X @{}}
    \toprule
    \textbf{Diagnosis} & \textbf{Template} \\
    \midrule
    Diabetes
      & \textit{Def.}\; Chronic hyperglycemia due to insulin
        defects. \newline
        \textit{Cas.}\; Retinopathy, neuropathy, nephropathy. \\
    \addlinespace
    Sepsis
      & \textit{Def.}\; Life-threatening organ dysfunction from
        dysregulated host response. \newline
        \textit{Cas.}\; Shock $\to$ AKI $\to$ respiratory
        failure. \\
    \addlinespace
    Heart failure
      & \textit{Def.}\; Inadequate cardiac output due to
        structural or functional impairment. \newline
        \textit{Cas.}\; Pulmonary edema $\to$ renal
        hypoperfusion. \\
    \bottomrule
  \end{tabularx}
  \end{apptable}
\end{table}

\subsection{Pool Diagnostics}
\label{app:pool_diagnostics}

\begin{table}[htbp]
  \centering
  \begin{apptable}
  \setlength{\tabcolsep}{3.5pt}
  \renewcommand{\arraystretch}{0.96}
  \caption{Knowledge-pool diagnostics (offline). Candidate support is measured over distilled links, and low-support attrition over concepts.}
  \label{tab:prompt_pool_quality}
  \begin{tabularx}{\linewidth}{Xcc}
    \toprule
    \textbf{Metric} & \textbf{III} & \textbf{IV} \\
    \midrule
    \multicolumn{3}{l}{\textit{Generation}} \\
    \# Unique ICD codes & 4{,}137 & 8{,}507 \\
    Format pass rate (regex/schema) & 97.6\% & 97.1\% \\
    Mean tokens / request & 148 & 156 \\
    \addlinespace
    \multicolumn{3}{l}{\textit{Grounding}} \\
    Ontology mapping success & 95.3\% & 93.8\% \\
    Ext.\ supported candidates (links, pre-filter) & 83.7\% & 81.9\% \\
    \addlinespace
    \multicolumn{3}{l}{\textit{Filtering diagnostics}} \\
    Format violation & 2.4\% & 2.9\% \\
    Failed mapping / missing concept & 4.7\% & 6.2\% \\
    Low external support (concepts) & 11.6\% & 12.8\% \\
    \addlinespace
    \multicolumn{3}{l}{\textit{After clustering}} \\
    \# Templates $M$ & 920 & 1{,}180 \\
    Avg.\ nodes / edges per template & 7.6 / 16.8 & 8.1 / 17.9 \\
    Median intra-cluster cosine dist. & 0.14 & 0.15 \\
    \bottomrule
  \end{tabularx}
  \end{apptable}
\end{table}

The externally supported rates in Table~\ref{tab:prompt_pool_quality} (83.7\% and 81.9\%) are pre-filter candidate pass rates over all distilled links, not support rates for the final pool.
Candidates without external support are removed before template construction, so every retained Hard-Import relation has grounded endpoints and support from PrimeKG or CCS.

The largest source of attrition is low external support (11--13\% of concepts, Table~\ref{tab:prompt_pool_quality}), where the LLM produces clinically plausible but ontologically unattested relations that the strict support filter correctly rejects.
After clustering, the resulting templates are compact (7--8 nodes, 17--18 edges on average), which keeps Hard Import efficient at inference time.

For \modelname{}\textsubscript{UMLS}, we replace PrimeKG with disease-complication and disease-finding relations from UMLS Metathesaurus~\cite{bodenreider2004unified}.
The vocabulary bridge, distillation prompt, grounding thresholds, and clustering procedure remain identical.
The resulting pool contains 874 (III) and 1{,}092 (IV) templates with comparable subgraph density (7.2 / 15.3 avg.\ nodes/edges).

\subsection{Clinical Audit}
\label{app:audit}

Structural support does not by itself establish clinical correctness, so we audited a frequency-stratified sample of 240 retained templates, 120 mapped to common diagnoses ($>$500 training occurrences) and 120 to rare ones ($<$5).
Two physicians independently reviewed each template, and disagreements were resolved by discussion.
Annotators viewed each template blinded to provenance and performance information and rated whether the definition is correct, whether each cascade relation is clinically correct, whether the cascade is adequate for the concept, and whether the template contains a major error that could mislead prediction.
For relation correctness, a template is counted as correct only when all of its cascade relations are judged clinically correct. Each percentage in Table~\ref{tab:audit} uses the number of audited templates in the corresponding stratum as its denominator.
Table~\ref{tab:audit} reports the results, and the higher error rates for rare diagnoses are discussed in the Limitations.

\begin{table}[htbp]
  \centering
  \begin{apptable}
  \setlength{\tabcolsep}{4pt}
  \caption{Template-level clinical audit of retained templates.}
  \label{tab:audit}
  \resizebox{\columnwidth}{!}{%
  \begin{tabular}{lccccc}
    \toprule
    \textbf{Stratum} & \textbf{$n$} & \textbf{Def.\ corr.} & \textbf{Rel.\ corr.} & \textbf{Casc.\ adeq.} & \textbf{Major err.} \\
    \midrule
    Common ($>$500) & 120 & 96.7\% & 96.7\% & 94.2\% & 0.8\% \\
    Rare ($<$5)     & 120 & 91.7\% & 90.8\% & 86.7\% & 2.5\% \\
    Overall         & 240 & 94.2\% & 93.8\% & 90.4\% & 1.7\% \\
    \bottomrule
  \end{tabular}}
  \end{apptable}
\end{table}

\subsection{Generator Dependence}
\label{app:generator}

To test dependence on a single proprietary generator, we rebuild the pool with Qwen3-32B under the identical prompt, grounding, filtering, and clustering procedure.
The two pools have similar candidate support and retain largely overlapping grounded links. Downstream AUPRC changes little across generators (Table~\ref{tab:generator}).

\begin{table}[htbp]
  \centering
  \begin{apptable}
  \setlength{\tabcolsep}{4pt}
  \caption{Generator dependence.}
  \label{tab:generator}
  \resizebox{\columnwidth}{!}{%
  \begin{tabular}{lccc}
    \toprule
    \textbf{Generator} & \textbf{Cand.\ support (III/IV)} & \textbf{Link overlap (III/IV)} & \textbf{AUPRC (III/IV)} \\
    \midrule
    GPT-4o    & 83.7\,/\,81.9\% & 100\,/\,100\% & 34.18\,/\,35.22 \\
    Qwen3-32B & 81.1\,/\,79.4\% & 76.8\,/\,74.9\% & 33.85\,/\,34.77 \\
    \bottomrule
  \end{tabular}}
  \end{apptable}
\end{table}

\section{Sensitivity}
\label{app:sensitivity}

\begin{figure*}[t]
  \centering
  \includegraphics[width=\linewidth]{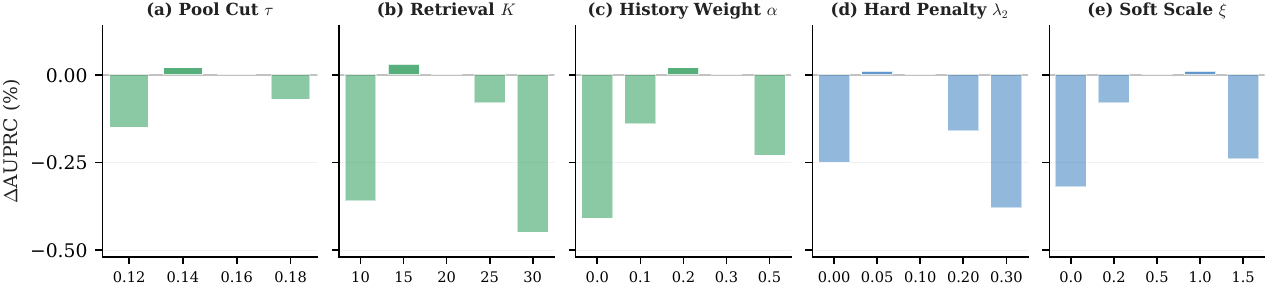}
  \caption{Sensitivity sweeps on MIMIC-IV. $\Delta$AUPRC is relative to the full-model reference in Table~\ref{tab:main_results}.}
  \label{fig:sensitivity}
\end{figure*}

Figure~\ref{fig:sensitivity} shows limited variation at moderate settings and larger losses at the extremes.

\paragraph{Pool and retrieval.}
The pool cut $\tau$ (panel~a) has minimal impact within the tested range, with less than 0.2 AUPRC variation.
The retrieval size $K$ (panel~b) shows an inverted-U pattern, where too few candidates ($K{=}10$) miss relevant templates and too many ($K{=}30$) dilute quality with noise.
The history weight $\alpha$ (panel~c) controls how much the retrieval scoring relies on trajectory context versus code-level similarity (Eq.~\ref{Eq:score}).
Moderate values ($\alpha \in [0.2, 0.3]$) perform best, while higher values over-weight trajectory context at the expense of visit-specific code matching.

\paragraph{Augmentation controls.} The Hard Import penalty $\lambda_2$ (panel~d) degrades performance at both extremes. High penalties suppress useful topology edits, while low penalties encourage excessive grafting. The default $\lambda_2{=}0.10$ lies in the stable range. The Soft Import scale $\xi$ (panel~e) is similarly stable at moderate values, with losses when feature editing is too strong ($\xi{>}1.0$) or absent ($\xi{=}0$).

\paragraph{Budget and deployment.} Larger per-visit budgets add latency and training cost with little AUPRC gain (Table~\ref{tab:topm_tradeoff}). The default $m{=}1$ offers the best accuracy-efficiency trade-off.

\begin{table}[htbp]
  \centering
  \begin{apptable}
  \setlength{\tabcolsep}{4pt}
  \caption{Per-visit budget $m$ trade-offs on MIMIC-IV.}
  \label{tab:topm_tradeoff}
  \begin{tabular}{lccc}
    \toprule
    \textbf{Setting} &
    \textbf{$\Delta$AUPRC} &
    \textbf{$\Delta$Lat.\ (ms)} &
    \textbf{$\Delta$Train (h)} \\
    \midrule
    \(m{=}1\) (default) & 0.00  & 0.0  & 0.0  \\
    \(m{=}2\)           & +0.03 & +4.5 & +3.9 \\
    \(m{=}3\)           & +0.01 & +8.7 & +9.0 \\
    \bottomrule
  \end{tabular}
  \end{apptable}
\end{table}

\section{Extended Ablation Results}
\label{app:extended_ablation}

This appendix provides additional component ablations and selector comparisons.
The rate-matched shuffle keeps the Stage~1 checkpoint, Stage~2 updates, training budget, and action rates fixed and shuffles only which visits receive each action, so it omits the policy forward pass and runs 1.4\,ms below the full model at the same skip rate (Table~\ref{tab:latency_breakdown}).
Always augment retrains Stage~2 without Skip and matches the Skip ablation in Figure~\ref{fig:ablation}.
In Table~\ref{tab:component}, the lower block reports MIMIC-IV deltas only, and removing the uncertainty signal $u_t$ also cuts the skip rate from 31\% to 12\%.
In Table~\ref{tab:selectors}, Labels identifies access to action-selection labels. The supervised selector uses training-time labels, while the oracle uses test outcomes. Eval./visit counts one decision pass for each learned selector or the heuristic, and candidate-loss evaluations for the oracle. The rate-matched shuffle requires neither. Always augment still makes one policy pass to choose among the $2K$ template-mode actions.
In Table~\ref{tab:overrides}, template overrides are measured on non-Skip visits, total departures include Skip, and positive-gain rates are post-hoc diagnostics rather than policy inputs. Within each replicate, the departure rate is $D=S+(1-S)O$, where $S$ is the visit-level Skip rate and $O$ the template-override rate among non-Skip visits. We compute each rate from the same visit records before averaging across replicates.
In Table~\ref{tab:richer}, both methods receive medication and procedure codes from the observed visits $V_{1:t}$ when predicting diagnoses at $V_{t+1}$.

\begin{table}[htbp]
  \centering
  \begin{apptable}
  \setlength{\tabcolsep}{4pt}
  \caption{Component ablations, AUPRC ($\Delta$).}
  \label{tab:component}
  \begin{tabular}{lcc}
    \toprule
    \textbf{Variant} & \textbf{MIMIC-III} & \textbf{MIMIC-IV} \\
    \midrule
    Full \modelname{} & 34.18 (0.00) & 35.22 (0.00) \\
    w/o refined pool & 31.51 ($-$2.67) & 32.07 ($-$3.16) \\
    Rate-matched shuffle & 33.09 ($-$1.09) & 33.82 ($-$1.40) \\
    w/o decoupled fusion & 33.39 ($-$0.79) & 34.31 ($-$0.92) \\
    \midrule
Frozen encoder & --- & $-0.53$ \\
w/o uncertainty signal $u_t$ & --- & $-0.48$ \\
w/o adaptive cascade depth & --- & $-0.45$ \\
w/o template credit & --- & $-0.37$ \\
Gate-only fusion & --- & $-0.26$ \\
    \bottomrule
  \end{tabular}
  \end{apptable}
\end{table}

\begin{table}[htbp]
  \centering
  \begin{apptable}
  \setlength{\tabcolsep}{4pt}
  \caption{Selector comparison and matched controls on MIMIC-IV.}
  \label{tab:selectors}
  \resizebox{\columnwidth}{!}{%
  \begin{tabular}{lccrc}
    \toprule
    \textbf{Selector} & \textbf{Labels} & \textbf{Eval./visit} & \textbf{AUPRC ($\Delta$)} & \textbf{Lat.\ (ms)} \\
    \midrule
    Uncertainty heuristic & None & 1 & 33.72 ($-$1.51) & 10.6 \\
    Supervised selector & Exhaustive & 1 & 34.53 ($-$0.69) & 11.7 \\
    Greedy myopic oracle & Test outcome & 41 & 34.80 ($-$0.43) & --- \\
    Contextual bandit & None & 1 & 34.69 ($-$0.53) & 11.5 \\
    \midrule
    Rate-matched shuffle & None & 0 & 33.82 ($-$1.40) & 10.4 \\
    Always augment & None & 1 & 34.59 ($-$0.64) & 12.8 \\
    \midrule
    Full RL & None & 1 & \textbf{35.22} & 11.8 \\
    \bottomrule
  \end{tabular}}
  \end{apptable}
\end{table}

\begin{table}[htbp]
  \centering
  \begin{apptable}
  \setlength{\tabcolsep}{4pt}
  \caption{Policy departures from code-only Top-1 retrieval.}
  \label{tab:overrides}
  \resizebox{\columnwidth}{!}{%
  \begin{tabular}{lcccc}
    \toprule
    \textbf{Dataset} & \textbf{Templ.\ ovr.} & \textbf{Total dep.} & \textbf{Pos.-gain} & \textbf{Forced Top-1} \\
    \midrule
    MIMIC-III & 31.9\% & 50.2\% & 67.1\% & 33.74 ($-$0.44) \\
    MIMIC-IV & 35.3\% & 55.5\% & 70.5\% & 34.64 ($-$0.58) \\
    \bottomrule
  \end{tabular}}
  \end{apptable}
\end{table}

\begin{table}[htbp]
  \centering
  \begin{apptable}
  \setlength{\tabcolsep}{4pt}
  \caption{Richer structured inputs on MIMIC-IV.}
  \label{tab:richer}
  \resizebox{\columnwidth}{!}{%
  \begin{tabular}{llccc}
    \toprule
    \textbf{Input} & \textbf{Method} & \textbf{AUPRC} & \textbf{Lat.\ (ms)} & \textbf{Skip/Soft/Hard} \\
    \midrule
    Diagnosis & KARE & 33.14 & 15.1 & --- \\
    Diagnosis & \modelname{} & 35.22 & 11.8 & 31.2/34.8/34.0\% \\
    + Med./proc. & KARE & 34.25 & 16.3 & --- \\
    + Med./proc. & \modelname{} & 36.20 & 12.9 & 35.7/39.7/24.6\% \\
    \bottomrule
  \end{tabular}}
  \end{apptable}
\end{table}

\section{Additional Results}
\label{app:additional_results}

\subsection{Label-Frequency Stratification}
\label{app:label_freq}

AUPRC is averaged over labels within each stratum, then over five seeds.
Performance declines toward the tail for all methods, while \modelname{}'s advantage over KARE generally grows (Table~\ref{tab:label_freq}).

\begin{table}[htbp]
  \centering
  \begin{apptable}
  \setlength{\tabcolsep}{2pt}
  \renewcommand{\arraystretch}{0.98}
  \caption{Per-stratum macro-AUPRC (\%). $\Delta$ is the gain over KARE.}
  \label{tab:label_freq}
  \begin{tabular}{l@{\hskip 5pt}cccccc}
    \toprule
    & \textbf{$>$500} & \textbf{200\,--} & \textbf{50\,--} & \textbf{20\,--} & \textbf{5\,--} & \textbf{$<$5} \\[-1pt]
    & & \textbf{500} & \textbf{200} & \textbf{50} & \textbf{20} & \\
    \midrule
    Trans.      & 33.48 & 25.24 & 16.88 & 12.58 & 10.03 & 5.95 \\
    HAP         & 36.15 & 27.65 & 19.02 & 14.38 & 11.22 & 7.06 \\
    SeqCare     & 34.25 & 29.37 & 22.28 & 17.03 & 12.59 & 6.99 \\
    KARE        & 36.46 & 31.44 & 24.28 & 18.57 & 14.58 & 8.45 \\
    \modelname{}& \textbf{38.45} & \textbf{32.94} & \textbf{26.94} & \textbf{22.15} & \textbf{17.55} & \textbf{11.97} \\
    \midrule
    $\Delta$    & +1.99 & +1.50 & +2.66 & +3.57 & +2.97 & +3.52 \\
    \bottomrule
  \end{tabular}
  \end{apptable}
\end{table}

\subsection{Calibration}
\label{app:calibration}

\begin{table}[htbp]
  \centering
  \begin{apptable}
  \setlength{\tabcolsep}{4pt}
  \caption{ECE on MIMIC-IV (mean$\pm$std over 5 seeds, macro-averaged over CCS labels). Lower is better.}
  \label{tab:calibration}
  \begin{tabular}{lc}
    \toprule
    \textbf{Method} & \textbf{ECE (\%) $\downarrow$} \\
    \midrule
    Transformer  & 8.12{\scriptsize$\pm$0.12} \\
    HAP          & 7.37{\scriptsize$\pm$0.09} \\
    SeqCare      & 6.95{\scriptsize$\pm$0.10} \\
    KARE         & 5.51{\scriptsize$\pm$0.08} \\
    \modelname{} & 4.76{\scriptsize$\pm$0.08} \\
    \midrule
    $\Delta$ vs.\ KARE & $-0.75$ \\
    \bottomrule
  \end{tabular}
  \end{apptable}
\end{table}

ECE uses 15 equal-width probability bins per label and averages across labels. \modelname{} has the lowest ECE among the compared methods on MIMIC-IV (Table~\ref{tab:calibration}).

\section{Mechanism Analysis}
\label{sec:appendix_c}

\subsection{Policy Behavior}
\label{sec:appendix_c_policy}

\begin{figure}[htbp]
    \centering
    \includegraphics[width=\linewidth]{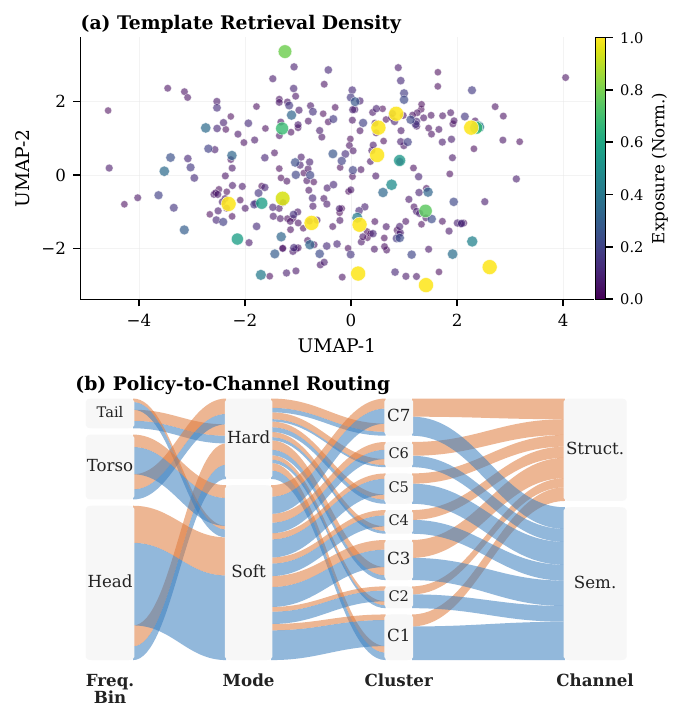}
    \caption{Template retrieval density (a) and policy-to-channel routing (b). }
    \label{fig:case_pool}
\end{figure}

Figure~\ref{fig:case_pool} shows template retrieval density (panel~a) and the routing structure linking import modes, template clusters, and representation channels (panel~b). Retrieval concentrates near cluster centers, while peripheral templates are rarely selected. Tail diagnoses favor Hard Import over Soft Import (Figure~\ref{fig:case_study}d).

\subsection{Reward Comparison}
\label{sec:appendix_c_paired}

\begin{table}[htbp]
  \centering
  \begin{apptable}
  \caption{Reward comparison on MIMIC-IV. Skip rates are reported as means.}
  \label{tab:paired_reward}
  \begin{tabular}{lcc}
    \toprule
    \textbf{Reward} & \textbf{AUPRC (\%)} & \textbf{Skip (\%)} \\
    \midrule
    Paired (default)                    & 35.22\scriptsize$\pm$0.41 & 31.2 \\
    Cross-entropy only & $34.10\pm0.39$ & 18.8 \\
    \bottomrule
  \end{tabular}
  \end{apptable}
\end{table}

The paired reward achieves higher AUPRC and a higher Skip rate than the cross-entropy-only reward. It measures the improvement from augmentation and penalizes Hard Import through the structural cost term.

\subsection{Qualitative Analysis}
\label{sec:appendix_c_qualitative}

\paragraph{Structural shortcut.} At $V_5$, Hard Import adds cascade nodes (Systemic Inflammation, Septic Shock) that create short message-passing paths within the visit graph (Figure~\ref{fig:topology_viz}). Across-visit attention combines this structural information with earlier chronic context.

\begin{figure}[htbp]
    \centering
    \includegraphics[width=\linewidth]{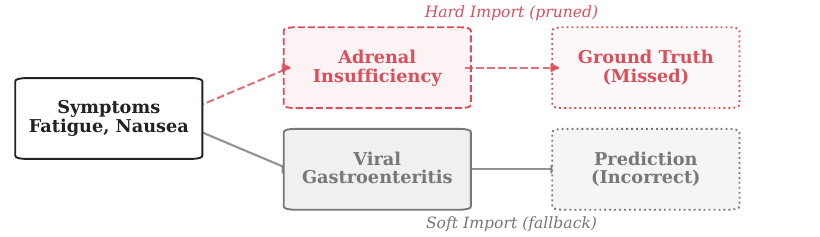}
    \caption{Over-pruning under ambiguity. The dashed path denotes a clinically plausible dependency pruned by the policy.}
    \label{fig:failure_case}
\end{figure}

\paragraph{Failure mode.} In Figure~\ref{fig:failure_case}, fatigue and nausea are compatible with both viral gastroenteritis and adrenal insufficiency. The Hard Import penalty $\lambda_2$ discourages grafting, so the policy selects Soft Import with the semantically closer viral gastroenteritis template, missing adrenal insufficiency. If augmentation barely changes prediction loss, the paired reward provides little corrective signal even for an incorrect prediction.

\end{document}